\documentclass[conference]{IEEEtran}
\IEEEoverridecommandlockouts

\usepackage{cite}
\usepackage{amsmath,amssymb,amsfonts}
\usepackage{algorithmic}
\usepackage{graphicx}
\usepackage{textcomp}
\usepackage{xcolor}
\usepackage{hyperref}

\def\BibTeX{{\rm B\kern-.05em{\sc i\kern-.025em b}\kern-.08em
    T\kern-.1667em\lower.7ex\hbox{E}\kern-.125emX}}
\begin{document}

\title{An Analysis of Artificial Intelligence Adoption in NIH-Funded Research}

\author{\IEEEauthorblockN{1\textsuperscript{st} Navapat Nananukul}
\IEEEauthorblockA{\textit{Information Sciences Institute} \\
\textit{University of Southern California}\\
Los Angeles, United States of America \\
nananuku@isi.edu}
\and
\IEEEauthorblockN{2\textsuperscript{nd} Mayank Kejriwal}
\IEEEauthorblockA{\textit{Information Sciences Institute} \\
\textit{University of Southern California}\\
Los Angeles, United States of America \\
kejriwal@isi.edu}
}

\maketitle

\begin{abstract}
Understanding the landscape of artificial intelligence (AI) and machine learning (ML) adoption across the National Institutes of Health (NIH) portfolio is critical for research funding strategy, institutional planning, and health policy. The advent of large language models (LLMs) has fundamentally transformed research landscape analysis, enabling researchers to perform large-scale semantic extraction from thousands of unstructured research documents. In this paper, we illustrate a human-in-the-loop research methodology for LLMs to automatically classify and summarize research descriptions at scale. Using our methodology, we present a comprehensive analysis of 58,746 NIH-funded biomedical research projects from 2025. We show that: (1) AI constitutes 15.9\% of the NIH portfolio with a 13.4\% funding premium, concentrated in discovery, prediction, and data integration across disease domains; (2) a critical research-to-deployment gap exists, with 79\% of AI projects remaining in research/development stages while only 14.7\% engage in clinical deployment or implementation; and (3) health disparities research is severely underrepresented at just 5.7\% of AI-funded work despite its importance to NIH's equity mission. These findings establish a framework for evidence-based policy interventions to align the NIH AI portfolio with health equity goals and strategic research priorities.
\end{abstract}

\begin{IEEEkeywords}
artificial intelligence, large language models, NIH funding, research portfolio analysis, health policy, health disparities, clinical implementation
\end{IEEEkeywords}

\section{Introduction}

Artificial intelligence (AI) and machine learning (ML) are transforming biomedical research, with applications spanning drug discovery, medical imaging, clinical decision support, and precision medicine \cite{chakraborty2023artificial,nananukul2024multi,wang2024large,guan2024drug}. The National Institutes of Health (NIH), as the primary funder of biomedical research in the United States, plays a critical role in setting the research agenda through its funding portfolio. Understanding where, how, and to what extent AI/ML technologies are being adopted across the NIH portfolio is essential for multiple stakeholders: NIH program officers and strategic planners must allocate resources to research areas with high potential impact; institutional research offices require insights into funding trends and competitive landscapes; and policymakers need evidence to inform science policy and digital health strategy.

Despite the rapid growth of AI/ML in biomedicine, no existing study has characterized AI adoption across the full scope of NIH-funded research at scale. Prior work has examined AI applications in specific disease areas (cancer imaging, neurodegenerative disease), individual funding mechanisms (SBIR/STTR), or narrowly defined technology domains (deep learning). While foundational ML approaches such as support vector machines and naive Bayes have enabled biomedical document analysis \cite{joachims1998text,mccallum1998comparison}, and curated datasets have catalyzed deep learning advances \cite{dernoncourt2017pubmed,baker2015automatic,chen2020litcovid}, these studies lack the breadth and granularity needed to answer critical policy questions at the portfolio level, a sample of which includes: what proportion of the NIH portfolio includes AI/ML research? Which disease areas receive the most AI funding? Are there disparities in AI adoption across institution types or research focuses? What is the relationship between AI-focused research and real-world clinical implementation?

The distribution of AI funding across disease areas has direct implications for health equity. The NIH has made health equity and health disparities a centerpiece of its strategic agenda, yet little is known about whether AI investment patterns align with these commitments \cite{eweje2022translatability,le2026national}. Infrastructure advantages in high-profile disease areas---large public datasets, established computational platforms, commercial partnerships---naturally create concentration in domains such as cancer, aging, and neuroscience. However, the converse is also true: health disparities research, minority health, rural health, and emerging infectious disease areas may lack the informatics infrastructure and computational tools to use recent AI advances. Understanding whether this gap reflects inevitable research maturity differences or a tractable structural mismatch is essential for NIH policy. Equitable AI adoption requires not just equalizing funding but ensuring that established computational methodologies are actively adapted and deployed to address health equity priorities, and that targeted investments in informatics infrastructure and clinical partnerships can narrow disparities.

Beyond funding concentration, another bottleneck exists between AI-focused research and real-world clinical deployment. Implementation science frameworks emphasize that translating innovations from research to practice requires deliberate strategy, sustained partnerships, and dedicated funding mechanisms \cite{proctor2022fast,armstrong2020welcome}. Early evidence suggests that the  majority of the NIH portfolio focuses on fundamental research, tool development, and methodological innovation, while comparatively fewer projects engage in the complex work of deploying AI systems into clinical settings, community health programs, or population health initiatives. This research-to-deployment gap is particularly acute in health disparities research, where limited informatics infrastructure combines with fewer established clinical AI partnerships to create compounding barriers. Additionally, workforce development in AI methods remains inadequate: the percentage of training grants that explicitly integrate AI/ML curricula is unknown, yet the ability to build a diverse, equity-focused AI research and implementation workforce is essential for long-term change. These three interrelated challenges i.e., disease-area disparities in AI infrastructure, the research-to-deployment translation gap, and workforce pipeline deficits, motivate systematic characterization of the current NIH AI portfolio.

To address these gaps, we conducted a large-scale computational analysis of 58,746 NIH-funded research projects, drawing upon large language models (LLMs) as automated coders to characterize the presence and nature of AI/ML research across the portfolio. Beyond portfolio prevalence and funding analyses, we also modeled university collaboration structure as a weighted network and identified community-level collaboration patterns using graph clustering.

\begin{itemize}
\item A human-in-the-loop LLM pipeline for large-scale classification of unstructured research abstracts into highly structured outputs that enable reproducible portfolio intelligence at scale.
\item Comprehensive empirical characterization of AI adoption across 58,746 projects in the NIH portfolio. We reveal that 1 out of 6 funded projects is related to AI. Moreover, AI projects receive a measurable 13.4\% funding premium relative to non-AI work. Analysis uncovers substantial disparities in AI investment concentration: cancer, aging, and mental health account for 50.1\% of all AI funding, while health disparities research—critical to NIH's equity mission—receives only 5.7\%, indicating a structural mismatch between stated priorities and actual funding patterns.
\item Identification of research-to-deployment gap: 79.0\% of AI projects remain in research and development stages, while only 14.7\% engage in clinical deployment or implementation, with health disparities research underrepresented at 5.7\% of AI-funded work.
\item A university collaboration network analysis that maps institutional coordination structure in NIH AI research (79 universities, 191 collaboration edges), and reveals collaboration communities anchored by a small set of high-intensity hubs and core institutions.
\item A network-science characterization of collaboration inequality, showing uneven collaboration capacity (heavy-tailed connectivity and concentrated betweenness) and a modular-but-bridge-mediated structure.
\end{itemize}

This work is critical for multiple stakeholders seeking to optimize the return on significant public research investments in digital health. By quantifying AI adoption patterns, funding disparities, and translation gaps at the portfolio level, we provide data-driven insights that enable NIH leadership and Congress to align funding priorities with stated commitments to health equity and strategic research impact. For research institutions, our analysis of collaboration networks and funding trends offers actionable intelligence for institutional planning and competitive positioning. For the broader biomedical research community, our findings highlight structural opportunities to strengthen the pipeline from AI-focused discovery to clinical deployment, particularly in underrepresented disease areas and health equity-focused domains. This work thus establishes a methodological and empirical foundation for more deliberate, evidence-informed stewardship of public research funding.

\begin{figure*}[h!]
\centering
\includegraphics[width=0.75\linewidth]{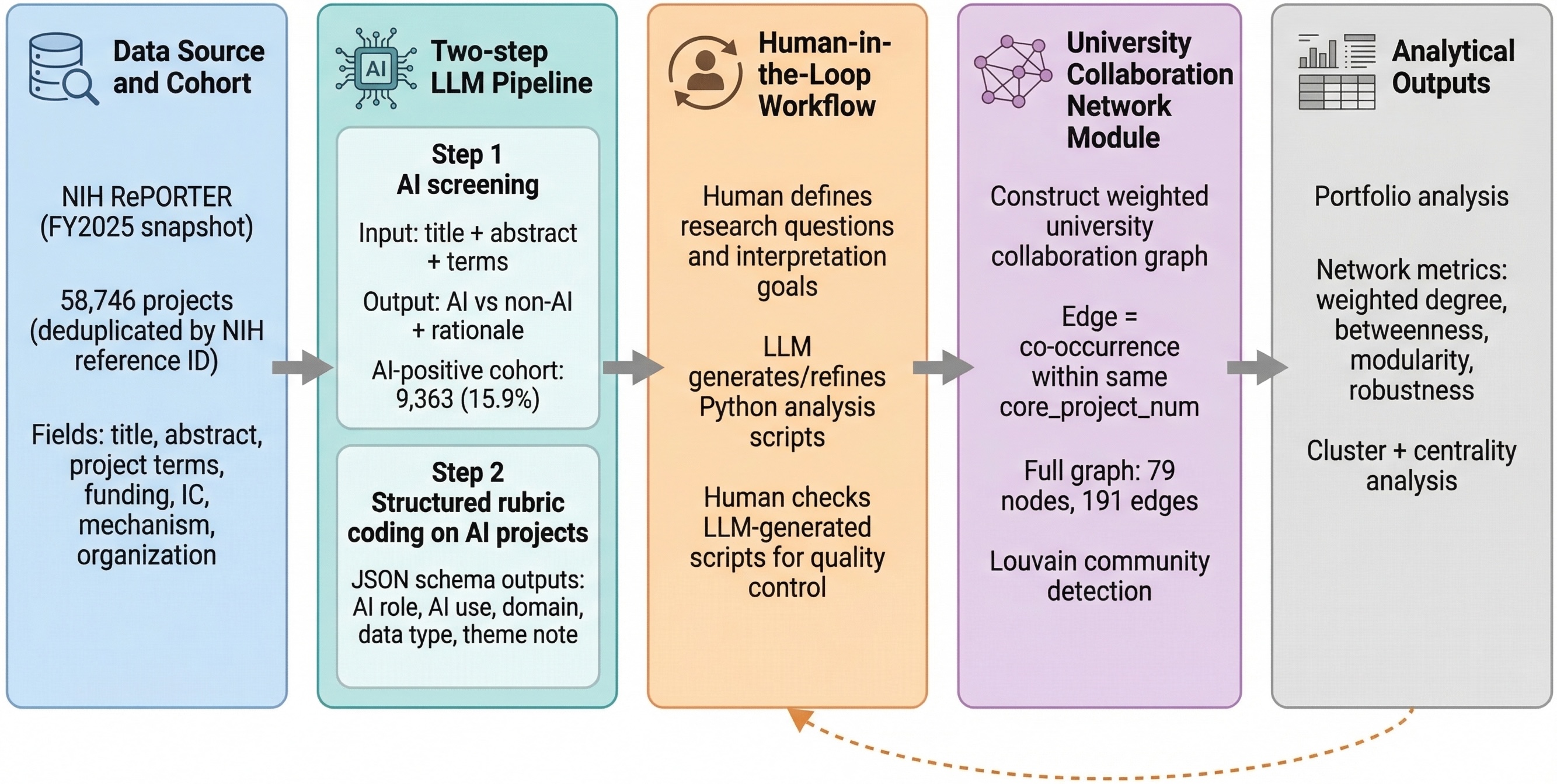}
\caption{Methodological workflow for AI portfolio analysis. The pipeline integrates: (a) data ingestion from NIH RePORTER (58,746 projects), (b) automated AI screening using large language models to classify projects as AI-positive or non-AI, (c) structured rubric coding on positive cases to extract role, use case, domain, and data type, (d) human-guided interpretation to refine research questions and validate LLM outputs, (e) network analysis of university collaborations with community detection, and (f) portfolio-level quantitative analysis of funding, clustering, and research trends.}
\label{fig:methods_overview}
\end{figure*}

\section{Related Work}

\subsection{LLMs in Scientific Research and Funding}

Integration of LLMs into the scientific workflow has transformed knowledge production and research planning. Recent studies have developed methodologies to quantify LLM usage in academic writing, observing rapid increases in scientific output across major publication platforms \cite{liang2025quantifying,kusumegi2025scientific}. The emergence of foundation models with substantial parameter counts and diverse training data has enabled breakthrough performance in few-shot and zero-shot learning \cite{brown2020language,chowdhery2023palm,touvron2023llama}. While LLMs enhance individual research creativity and proposal fluency \cite{doshi2024generative,gao2024quantifying}, evidence suggests these tools may narrow the collective focus of science by concentrating research directions \cite{hao2026artificial}. Beyond general-purpose models, domain-specific architectures in natural language processing \cite{phan2021scifive} and knowledge graph \cite{10.1007/978-3-032-09530-5_11} have been used to improve model performance. While recent applications demonstrate strong zero-shot classification performance on biomedical tasks such as pathology report coding \cite{sushil2024comparative} and sentiment analysis of clinical survey data \cite{lossioventura2024comparison}, recent work has found that adoption of promotional language in funding applications is positively associated with funding success \cite{peng2024promotional}. Our approach differs from these studies by using LLMs not as a researcher tool but as automated coders for large-scale semi-autonomous portfolio analysis, addressing the challenge of rapid knowledge synthesis across thousands of research abstracts.

\subsection{Implementation Science, Health Equity, and AI Bias}

Translating AI-driven innovations into healthcare requires robust implementation science (IS) frameworks. Recent work has proposed the FAST framework to assess speed of translation from discovery to policy, with emerging perspectives on rapid cycle research in cancer care delivery \cite{norton2023advancing}. However, researchers emphasize that health equity must be a primary consideration in implementation science to prevent the automation of historical biases \cite{brownson2021implementation,straw2020automation}. Ethical frameworks for machine learning in medicine underscore the need to address algorithmic fairness and transparency \cite{vayena2018machine}, particularly given documented evidence of racial bias in widely-deployed clinical risk prediction algorithms \cite{obermeyer2019dissecting}. Several studies have identified that without proactive monitoring and application of equity lenses like the PRISM framework \cite{fort2023applying,shelton2020extension}, AI models risk exacerbating existing health disparities \cite{zou2018ai,10.1001/jamanetworkopen.2025.12474}. Policy frameworks such as the EU AI Act are being developed to regulate AI systems and ensure safety through systematic evaluation \cite{eu2023aiact,gama2022implementation}. Our work contributes to this landscape by identifying specific disparities in AI adoption across disease areas and institutions within the NIH portfolio, providing empirical evidence for targeted equity interventions.

\subsection{Machine Learning for Scientific Text Classification}

Machine learning and NLP methods for document classification have evolved substantially over the past three decades. Early approaches using support vector machines \cite{joachims1998text}, naive Bayes classifiers \cite{mccallum1998comparison}, and decision trees \cite{lewis1994comparison} established foundational techniques for text categorization. These methods were subsequently applied to specialized biomedical tasks, including automated diagnosis coding from radiology reports \cite{karimi2017automatic} and suicide risk assessment from clinical notes \cite{bittar2019text}. To support continued advancement, curated benchmark datasets have become essential resources: the PubMed 200k RCT dataset for medical abstract sentence classification \cite{dernoncourt2017pubmed}, automatic coding of cancer hallmarks \cite{baker2015automatic}, and the LitCovid database for pandemic-related literature \cite{chen2020litcovid}. More recently, evaluations of data balance in biomedical document classification have highlighted challenges in handling unbalanced training sets common in healthcare applications \cite{laza2011evaluating}. Machine learning and NLP are increasingly applied to classify and analyze large-scale qualitative and scientific text data, often employing human-assisted approaches to synthesize narratives or identify patterns \cite{towler2022applying,criss2023solidarity}. These methods enable rapid processing of diverse datasets, from patient experiences in healthcare \cite{bastiaansen2022experiences} to heterogeneous clinical data \cite{palacios2020assessing}. However, maintaining the generalizability and transportability of ML models remains challenging, necessitating adaptive implementation strategies \cite{geng2023adaptation}. Our two-step LLM pipeline addresses these challenges through human-in-the-loop refinement and confidence calibration, demonstrating that structured human feedback can substantially improve model performance and interpretability when applied to biomedical abstracts at scale.

\subsection{Research Portfolio Analysis and Funding Landscape Studies}

Our work extends previous research on research portfolio analysis and funding landscape characterization. Research funding mapping plays a critical role in pandemic preparedness and global health security, with recent efforts to establish living mapping reviews of disease-specific research investments \cite{app152211853,seminog2024protocol} and surveillance networks for emerging infectious threats \cite{carroll2021preventing}. While prior work has examined specific aspects of the biomedical research landscape---such as AI applications in particular disease areas \cite{naskar2025biomedical,nananukul2026clinicbotguidelinegroundedclinicalchatbot,mathur2020artificial}, health policy implications of digital health, or workforce development challenges---no study has comprehensively characterized AI adoption across the entire NIH portfolio \cite{richardson2022framework,badr2024digital,abramoff2022reimbursement}. Portfolio-level analysis complements disease-specific studies by identifying cross-cutting themes, funding disparities, and policy-relevant patterns that would not emerge from examination of individual research areas. This enables policy makers to view the research landscape holistically and identify strategic opportunities for targeted investment.

\section{Methods}

Our analytical workflow comprises five integrated stages illustrated in Figure~\ref{fig:methods_overview}: (1) Data preparation: compilation and deduplication of NIH RePORTER project records for FY2025 (58,746 projects). (2) AI screening: automated identification of projects with substantive AI/ML involvement using LLM-based classification. (3) Structured rubric coding: extraction of analysis-ready labels for AI use, contribution type, domain, data type, and thematic notes on positive cases. (4) Human-guided validation: investigator review, refinement of recoding rules, and reanalysis to ensure accuracy and uncover hidden research patterns. (5) Portfolio analysis and network characterization: quantification of funding patterns, translational gaps, disease disparities, and institutional collaboration structure via network analysis. The pipeline is iterative, combining reproducible computational classification with expert validation for policy-relevant insights.

\subsection{Data Source and Cohort}

We analyzed NIH-funded projects using data from the NIH Research Performance and Reporting System (RePORTER)\footnote{\href{https://reporter.nih.gov}{https://reporter.nih.gov}}, accessed through publicly available project-level records. The study cohort was constructed as a single fiscal-year snapshot (FY2025) and comprised 58,746 projects after ID-level consolidation and quality checks. Each record was indexed by NIH reference ID and linked to structured project metadata.

For each project, we extracted fields used in downstream classification and portfolio analysis: project title, abstract text, project terms, total funding amount, administering institute/center (IC), funding mechanism, organization name, and organization type. These fields were selected because they jointly support (i) identification of substantive AI/ML involvement from narrative text, and (ii) policy-relevant stratification by disease focus, institution type, and funding patterns.

Records were deduplicated by NIH reference ID; text fields were normalized (whitespace/encoding cleanup); and missing values were retained as explicit null/unknown categories rather than dropped, to preserve denominator integrity in portfolio-level summaries. Project metadata were stored in line-delimited JSON format to support deterministic joins with AI classification and rubric outputs.

The primary unit of analysis was the individual project abstract, supplemented by title and project terms. We used abstracts as the core analytic text because they are standardized, available at scale, and provide sufficient methodological and application context for portfolio-level coding. Funding and institutional variables were analyzed as project-level attributes linked to each abstract-level classification.

\subsection{Two-step LLM Pipeline}

NIH RePORTER provides rich project-level metadata (title, abstract, project terms, administering IC, funding mechanism, organization type, and funding amount), but it does not provide a standardized label for whether AI/ML is a substantive method, how AI is used, or how projects map to policy-oriented AI categories. These questions are central to our study and cannot be answered reliably with simple keyword matching alone, because biomedical abstracts frequently use heterogeneous terminology and mixed methodological language (e.g., statistical modeling, bioinformatics, and AI terms in the same narrative).

To address this gap, we implemented a two-pass LLM pipeline that separates broad screening from detailed coding. This architecture allows portfolio-scale processing while preserving interpretable outputs for downstream policy analysis.

\textbf{Step 1: Portfolio-wide AI screening.}
All 58,746 projects were processed with a conservative zero-shot prompt using GPT-4o-mini. The model was instructed to classify a project as AI-relevant only when the abstract described a \emph{substantive} AI/ML role (development or application of AI/ML methods), rather than generic computational/statistical support. Inputs included title, abstract, and project terms. Outputs included a binary AI/non-AI label and model reasoning/confidence metadata for traceability. Pass 1 identified 9,363 AI projects (15.9\%), which formed the analysis cohort for rubric coding.

\textbf{Step 2: Structured rubric coding of AI projects.}
Only AI-positive projects from Pass 1 were sent to a fixed-schema rubric prompt. For each project, the model returned one JSON object with controlled fields:
\texttt{what\_ai\_used\_for}, \texttt{ai\_contribution}, \texttt{ai\_role},
\texttt{primary\_focus\_areas}, \texttt{application\_domain},
\texttt{application\_domain\_other\_specify}, \texttt{type\_of\_aiml},
\texttt{data\_type}, and \texttt{theme\_note}.
This step transformed unstructured narrative text into analysis-ready categorical outputs used for prevalence, co-occurrence, funding-by-category, and qualitative synthesis.

\textbf{Prompt constraints and output standardization.}
The rubric prompt enforced exact key names, closed category vocabularies, and JSON-only responses. It also required short free-text clarification when application\_domain = Other. These constraints reduced format drift and enabled deterministic aggregation across thousands of projects.

\subsection{Human-in-the-Loop Analytic Workflow}

A central methodological feature of this study is a human-in-the-loop workflow in which the human investigator defines analytical intent and policy questions, while the LLM agent implements executable analysis code and iteratively updates outputs. The interaction is not limited to prompting for narrative summaries; instead, it is used to produce and refine reproducible Python pipelines.

\textbf{Human role.} The investigator specified (i) research questions and prioritization criteria, (ii) interpretation goals for policy audiences, (iii) logic adjustments (e.g., handling of ``Other'' categories), and (iv) figure/storyline requirements for publication.

\textbf{LLM role.} The LLM agent translated these requirements into analysis scripts, performed data joins and aggregations, generated figures, and updated outputs after each human feedback cycle.

The resulting implementation comprised 7 Python scripts for data aggregation, cross-tabulation, co-occurrence analysis, and visualization. Execution produced 249 output files (tables, summaries, and figures), including 77 core analysis outputs and 167 generated figures. Iterative human review was used to refine decision rules and recoding logic (particularly for ambiguous ``Other'' categories), after which analyses were rerun to produce final reproducible results.

\begin{figure*}[h!]
\centering
\includegraphics[width=0.85\linewidth]{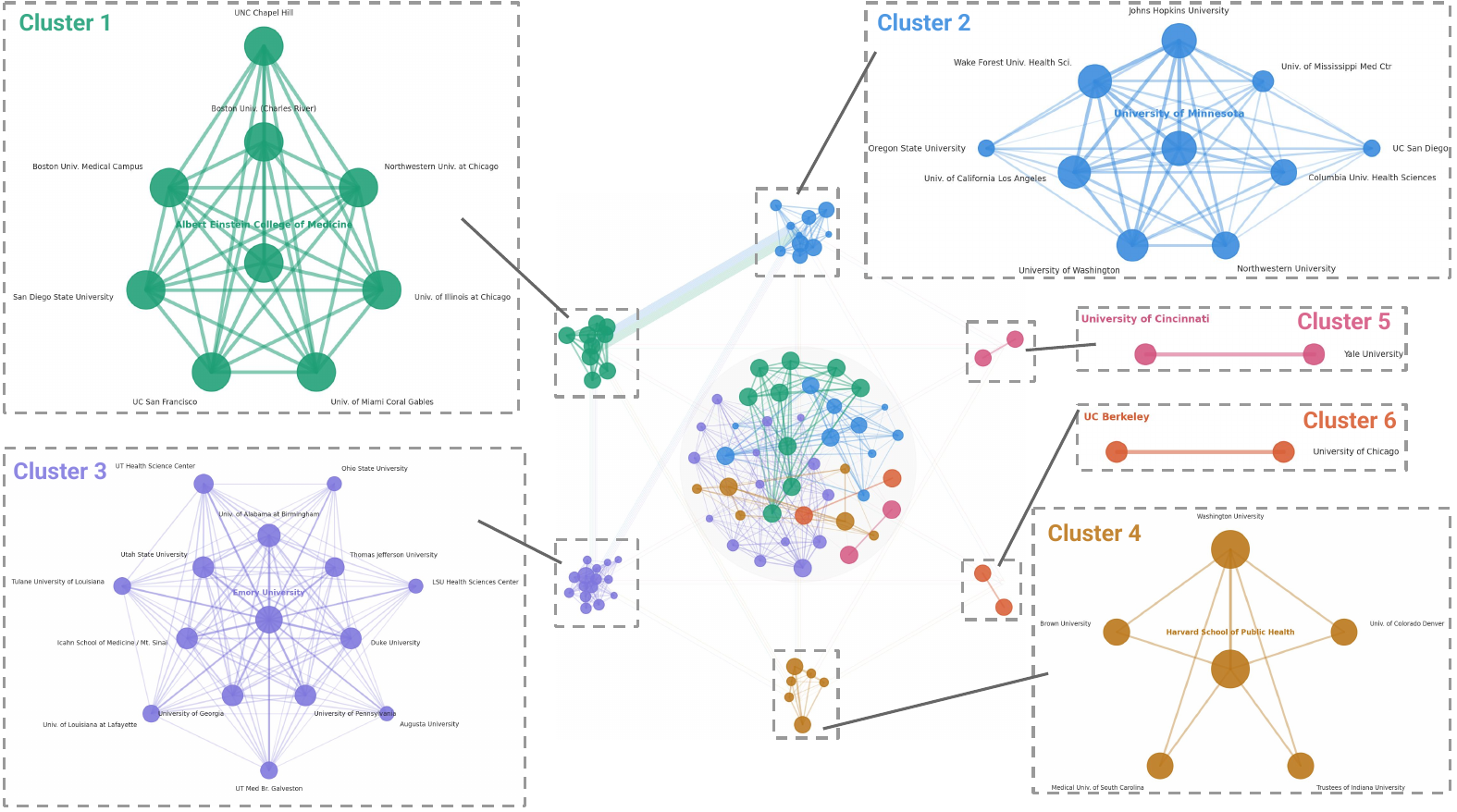}
\caption{University collaboration network from NIH AI projects showing six communities detected via Louvain modularity optimization on the largest connected component (48 nodes, 158 edges). Each node represents a university organization, with node size proportional to degree centrality (number of collaborations). Edges represent co-participation in project clusters, with thickness indicating collaboration frequency.}
\label{fig:collab_network}
\end{figure*}

\subsection{University Collaboration Network}
We illustrate the network and clusters in Figure~\ref{fig:collab_network}. We constructed an undirected, weighted university collaboration network from the full deduplicated NIH portfolio. Each node represents a university organization. Two universities are connected if they co-occur in the same project cluster. Edge weight equals the number of such co-occurrences across the dataset (collaboration count). We used this cluster-based co-participation definition because collaborator-level subaward splits are not available in the source metadata. At the full-network level, the graph contains 79 universities and 191 weighted edges (density 0.062), with 9 connected components and a largest connected component (LCC) of 48 nodes. We quantified node- and network-level structure using weighted degree (collaboration volume), betweenness centrality (bridge role), clustering coefficient (local cohesion), assortativity, modularity, and robustness under targeted hub removal.

We performed community detection with Louvain on the weighted collaboration network, using edge weights as collaboration counts and modularity optimization as the objective. For stable interpretation, we report communities on the LCC (48 nodes, 158 edges), where community structure is most informative and less affected by isolated mini-components.

\subsection{Reproducibility}  
The workflow is packaged as staged Python scripts with fixed input/output paths, making reproduction straightforward. It runs with Python 3.10+ and dependencies in requirements.txt. Core data inputs are: (i) project metadata (project\_details.jsonl), (ii) AI screening results (ai\_classification\_results.jsonl), and (iii) rubric coding results (rubric\_results.jsonl). Outputs are written to four analysis directories: Descriptive-analysis/, rubric-analysis/analysis/, follow-up-analysis/data/, and publication-findings/, as tables (CSV/JSON/Markdown) and figures (PNG).

Users can reproduce results in two modes: (1) \textit{full rerun}, including both LLM passes (AI screening + rubric coding), or (2) \textit{analysis-only rerun}, reusing provided Pass-1/Pass-2 outputs and regenerating all downstream tables/figures. Long-running stages are restartable via progress/state files, so interrupted runs continue without recomputing completed items. In our final execution, the pipeline produced 249 output files, including 77 core analysis artifacts and 167 figures.

\begin{figure*}[h!]
\centering
\includegraphics[width=0.8\linewidth]{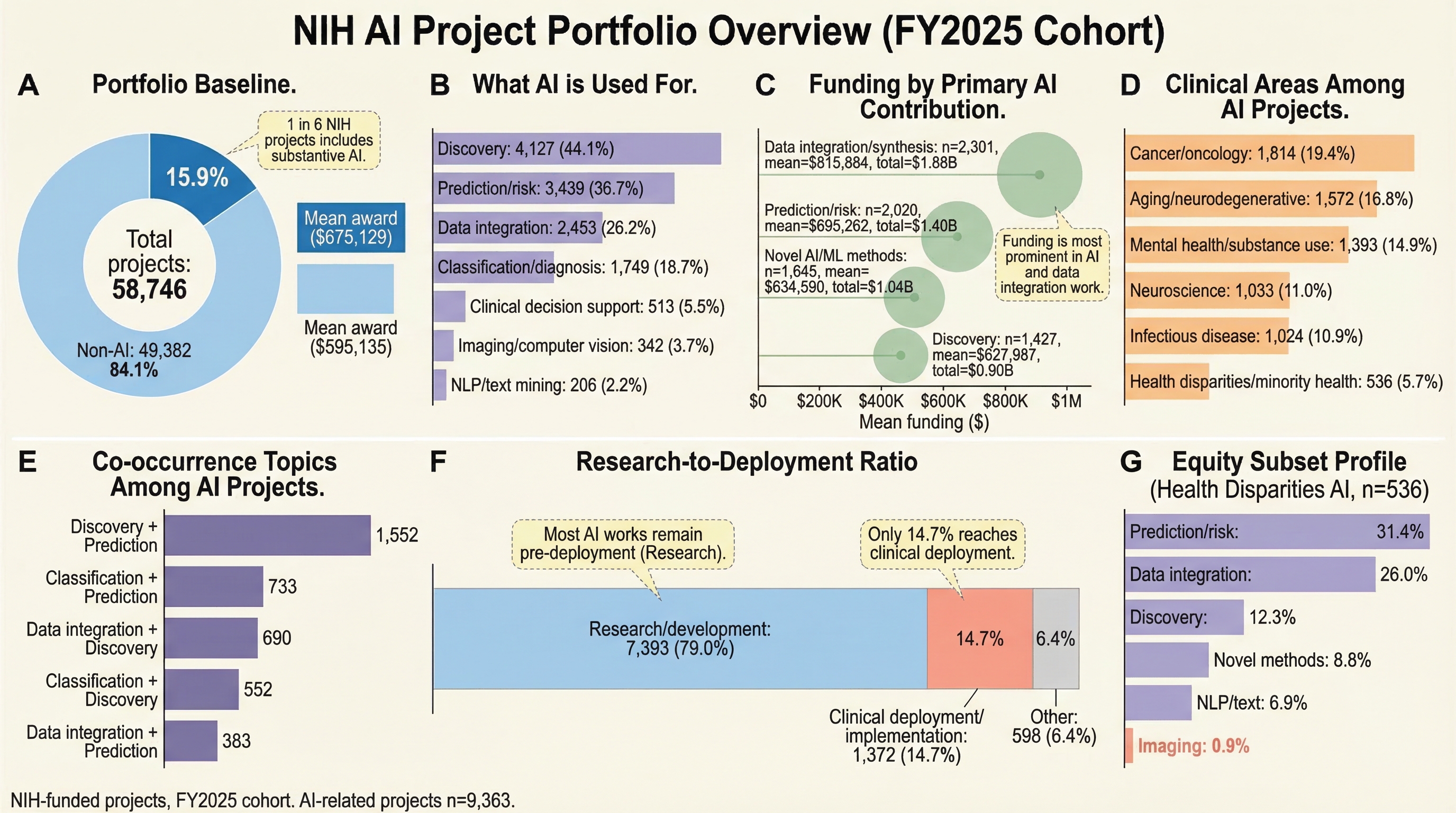}
\caption{Quantitative landscape of AI adoption across the NIH portfolio. Panel A: AI prevalence and funding premium across 58,746 projects. Panel B: Primary application categories (discovery, prediction, data integration). Panel C: Funding intensity by contribution type. Panel D: Disease area concentration. Panel E: Co-occurrence patterns of primary AI applications. Panel F: Distribution across research-to-deployment pipeline. Panel G: Application profile in health disparities-focused research.}
\label{fig:quantitative_landscape}
\end{figure*}

\section{Results}

Our analysis of the 58,746 NIH-funded projects from FY2025 reveals a multifaceted landscape of AI adoption that spans portfolio-level concentration patterns, institutional collaboration structures, and methodological sophistication. We organize our findings into three thematic areas: (1) the quantitative composition and funding dynamics of the AI portfolio, (2) the institutional collaboration networks that sustain AI research, and (3) the semantic characteristics and hidden complexity uncovered through human-guided LLM analysis.

\noindent\textbf{Finding 1: AI contributed to substantial funded research activity in drug/thoery discovery, prediction, and data integration.}

As shown in Figure \ref{fig:quantitative_landscape} (A), AI constituted 15.9\% of the NIH portfolio (9,363 of 58,746 projects), with AI-related projects receiving a mean award of \$675,129 compared to \$595,135 for non-AI projects, representing a 13.4\% funding premium. This establishes AI as a substantial and preferentially supported research modality. Figure \ref{fig:quantitative_landscape} (b) shows that the portfolio is dominated by three primary application categories: discovery research (4,127 projects, 44.1\%), prediction and risk assessment (3,439 projects, 36.7\%), and data integration/synthesis (2,453 projects, 26.2\%). Moreover, Figure \ref{fig:quantitative_landscape} (C) illustrates that among contribution types, funding intensity varies substantially, with data integration and synthesis projects (n=2,301) attracting the highest mean funding at \$815,884 per award for a total of \$1.88 billion, followed by prediction and risk assessment projects (n=2,020) at \$695,262 mean per award (total \$1.40 billion). Infrastructure-like AI work---especially data integration---attracts the largest funding intensity and represents a dominant research strategy in the NIH portfolio.

\noindent\textbf{Finding 2: AI work is concentrated in specific clinical topics and specialties.}

AI activity is not evenly distributed across disease domains. Figure \ref{fig:quantitative_landscape} (D) shows that Cancer and oncology lead with 1,814 projects (19.4\% of AI-funded work), followed by aging and neurodegenerative research (1,572 projects, 16.8\%), and mental health and substance use disorders (1,393 projects, 14.9\%). Neuroscience (1,033 projects, 11.0\%) and infectious disease (1,024 projects, 10.9\%) represent secondary foci. In contrast, health disparities and minority health research comprises only 536 projects (5.7\%), revealing substantial underrepresentation relative to these domains' importance in NIH's mission. Methodologically, Figure \ref{fig:quantitative_landscape} (E) shows co-occurrence between AI methods. Approximately 1,552 projects combine both discovery and prediction in their research design, suggesting that many NIH-funded studies integrate target or biomarker discovery with risk prediction within a single investigation. The next most common pairing is classification combined with prediction (733 projects), followed by data integration paired with discovery (690 projects). These co-occurrence patterns reflect a translational research logic where computational methods first identify novel biomarkers or targets and then validate their predictive utility.

\noindent\textbf{Finding 3: Most AI projects remain in research and development stage.}

Figure \ref{fig:quantitative_landscape} (F) illustrates that most AI-related projects are at research-stage: 79.0\% (7,393 projects) are classified as research or development stage, only 14.7\% (1,372 projects) are engaged in clinical deployment or implementation, and 6.4\% (598 projects) fall into other categories. This distribution indicates that the NIH portfolio remains heavily research-focused, with mechanisms to bring AI innovations into real-world clinical care remaining underdeveloped. This finding carries significant policy implications, suggesting that while the scientific and technical foundations for AI-enabled healthcare are expanding rapidly, the translation infrastructure remains limited.

\begin{figure*}[h!]
\centering
\includegraphics[width=0.85\linewidth]{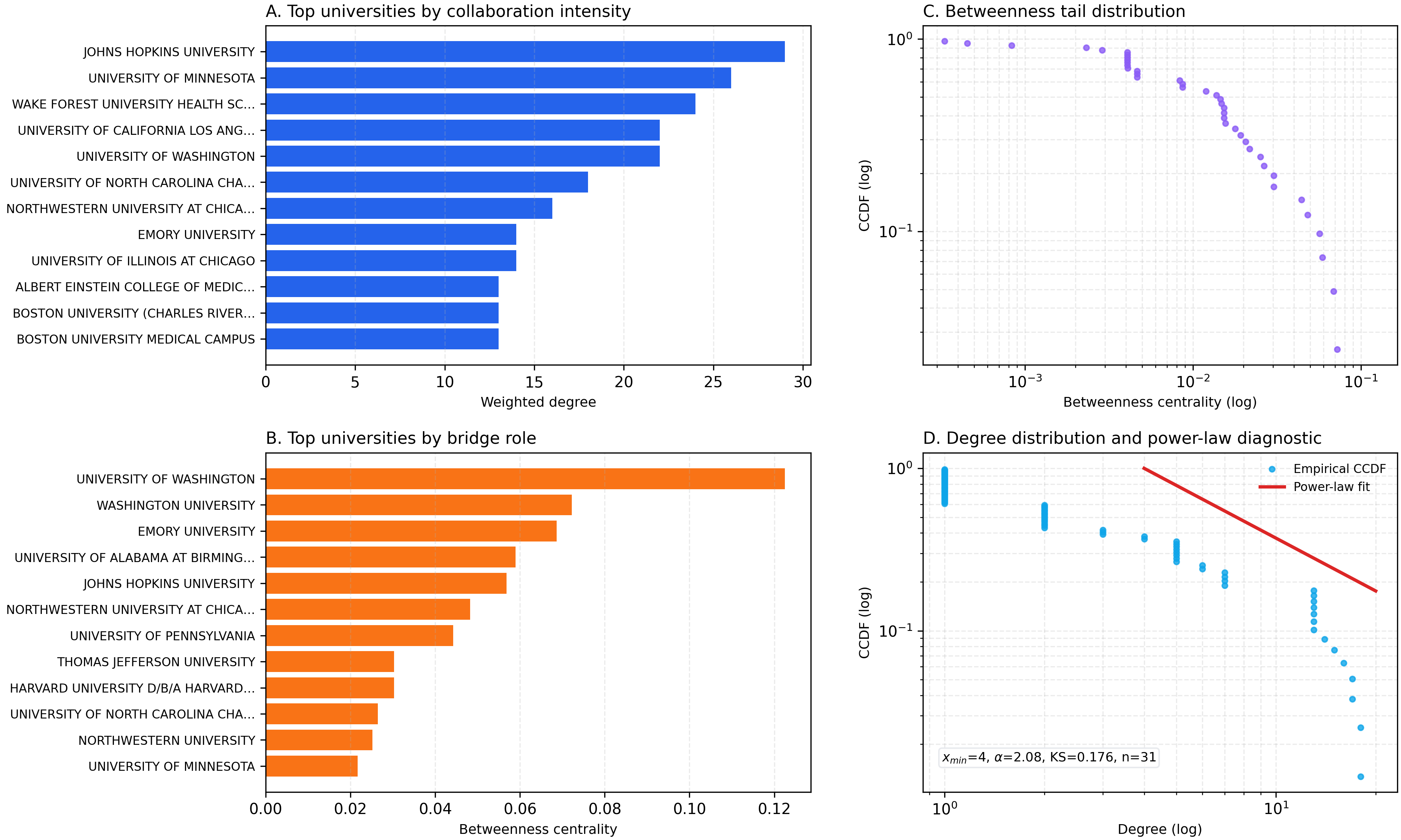}
\caption{University collaboration network centrality analysis and structure. (Panels A and B) Top universities ranked by complementary centrality measures: Panel A shows weighted degree centrality (collaboration intensity), with Johns Hopkins University leading, while Panel B shows betweenness centrality (bridge role), with University of Washington as the top connector. Network structure diagnostics: Panel C displays the betweenness centrality complementary cumulative distribution function (CCDF) on log-log scale, showing most institutions cluster near low betweenness with a small high-bridge subset. Panel D shows the degree distribution CCDF with power-law fitted tail.}
\label{fig:university_collaboration_top_hubs}
\end{figure*}

\noindent\textbf{Finding 4: Health equity research and AI has a distinct profile: prediction and data integration heavy}

Among the 536 health disparities-focused AI projects, the application profile diverges substantially from the overall portfolio as shown in Figure \ref{fig:quantitative_landscape} (G). Prediction and risk assessment account for 31.4\%, data integration for 26.0\%, and discovery for 12.3\%. Novel method development comprises 8.8\%, while natural language processing and text mining represent 6.9\%. Notably, imaging applications appear in only 0.9\% of disparities-focused AI work, indicating that computer vision methodologies remain largely absent from health equity research despite their prominence in the broader AI portfolio. This methodological gap suggests that high-throughput imaging, while central to cancer and aging research, is not yet integrated into the workflows of health disparities researchers, potentially limiting the application portfolio for equity-focused AI research.

\noindent\textbf{Finding 5: Clustering reveals university collaborations patterns.}

Clustering on the largest connected component of the university collaboration graph (48 nodes, 158 edges) identified six communities. In each community, we identified main anchor institutions within the cluster. In Cluster 1, a compact block is co-anchored by Albert Einstein College of Medicine, UCSF, Northwestern at Chicago, UIC, and Boston University. In Cluster 2, University of Minnesota, Johns Hopkins, and Wake Forest act as core anchors. In Cluster 3, Emory, UAB, and University of Pennsylvania form the core cluster. This pattern indicates that community structure is not arbitrary; it is organized around a small set of anchor universities that sustain dense local collaboration. A smaller but cohesive cluster (Cluster 4) includes Harvard School of Public Health and Washington University, plus Brown, MUSC, Indiana-affiliated, and Colorado Denver nodes. This cluster appears as a compact secondary collaboration pocket. Clusters 5 and 6 show compact peripheral links between University of Cincinnati--Yale University and UC Berkeley--University of Chicago, respectively.

\noindent\textbf{Finding 6: The hub and bridge roles are concentrated in specific universities, revealing how communities are held together.}

Figure~\ref{fig:university_collaboration_top_hubs} shows that the institutions with the highest collaboration volume are not identical to those with the strongest bridge function. On weighted degree (Panel A), \textit{Johns Hopkins University} ranks first (29), followed by \textit{University of Minnesota} (26), \textit{Wake Forest University Health Sciences} (24), and both \textit{UCLA} and \textit{University of Washington} (22). These institutions represent the most collaboration-intensive hubs within dense community cores. In contrast, betweenness centrality (Panel B) highlights bridge institutions that connect otherwise separated parts of the network. \textit{University of Washington} is the top bridge node (0.1225), followed by \textit{Washington University} (0.0723), \textit{Emory University} (0.0686), \textit{University of Alabama at Birmingham} (0.0590), and \textit{Johns Hopkins University} (0.0568). Notably, \textit{Washington University} has relatively low weighted degree (4) but very high betweenness, indicating a connector role rather than a high-volume local hub.

This hub--bridge separation aligns with the Louvain community pattern: high weighted-degree institutions (e.g., Johns Hopkins, Minnesota, Wake Forest) anchor dense intra-community collaboration, while high-betweenness institutions (especially University of Washington, Emory, UAB, and Washington University) provide cross-community integration. The network is therefore modular but coordinated through a small bridge layer.

\noindent\textbf{Finding 7: University collaboration capacity is uneven, with a small bridge layer connecting clusters.}

Figure~\ref{fig:university_collaboration_top_hubs} (Panel C and D) indicate that collaboration roles are strongly concentrated. Panel C (betweenness CCDF, log-log scale) shows that most universities cluster near very low betweenness, while only a small subset occupies high bridge-centrality positions, indicating that cross-community connectivity depends on relatively few connector institutions. Panel D (degree CCDF with fitted tail) is consistent with a heavy-tailed collaboration pattern ($x_{\min}=4$, $\alpha=2.08$, KS=0.176), where many universities have limited collaboration breadth and a minority account for disproportionately high connectivity. 

Together with the rankings in Panels A and B, this supports a structural interpretation of the network as \emph{modular but uneven}: dense local collaboration communities are sustained by high-volume hubs, while inter-community coordination relies on a narrow bridge layer. In practical terms, this implies that collaboration opportunities and network influence are not broadly distributed across institutions.

\begin{table}[h!]
\centering
\caption{Most Common Words and Phrases from Project Descriptions}
\label{tab:common_phrases}
\footnotesize
\setlength{\tabcolsep}{4pt}
\begin{tabular}{p{5.4cm}r}
\hline
\textbf{Phrase} & \textbf{Count} \\
\hline
Predictive modeling & 35 \\
Machine learning & 23 \\
AI for analyzing cognitive biological data & 19 \\
Deep learning & 16 \\
Predictive modeling for treatment outcomes & 13 \\
Computational modeling & 12 \\
Single cell & 10 \\
Computational modeling for drug discovery & 9 \\
Predictive modeling for individualized clinical trajectories & 9 \\
AI for automating radiotherapy treatment planning & 9 \\
Risk prediction & 8 \\
Bioinformatics for gene expression analysis & 8 \\
Dynamic risk calculators for individual patient outcomes & 8 \\
MA sequencing & 6 \\
Cells & 6 \\
Multi-omics & 5 \\
Drug discovery & 3 \\
\hline
\end{tabular}
\end{table}

\noindent\textbf{Finding 8: Project descriptions reveal prediction-focused and concrete clinical goals.}

Beyond the category-level statistics, examination of actual project language in Table~\ref{tab:common_phrases} reveals a highly interpretable research agenda. Detailed inspection of representative phrases shows clinically grounded AI applications rather than abstract methodological work: ``predictive modeling for treatment response'' (35 occurrences), ``AI for analyzing complex biological data'' (19), ``AI for analyzing single-cell RNA sequencing data'' (15), ``predictive modeling for treatment outcomes'' (13), ``single-cell RNA sequencing for gene expression analysis'' (11), and ``AI for automating radiotherapy treatment planning'' (9). These concrete examples demonstrate that projects articulate substantive AI applications aligned with biomedical objectives, confirming that the portfolio includes not just tool-building but clinically interpretable research aims with direct translational potential.

\noindent\textbf{Finding 9: Human-in-the-loop LLM analysis recovers hidden clinical domains, revealing underrepresented research areas.}

The initial automated rubric classification resulted in 41.5\% of projects classified as ``Other'' for disease area, indicating substantial ambiguity in standard categorical coding. Through structured human-in-the-loop refinement, we reduced the ``Other'' category to 17.7\%, discovering previously undetected disease domains now visible in Table~\ref{tab:hidden_domains}. Top recovered terms include alcohol use disorder (25 projects), chronic pain management (23), autoimmune diseases (23), autism spectrum disorder (21), ophthalmology (19), and kidney transplantation (18). This 23.8 percentage point improvement in classification clarity demonstrates that LLM-assisted analysis, coupled with expert validation, can uncover overlooked research areas and improve portfolio comprehensiveness. The recovery of behavioral health, pain management, and autoimmune disease areas---domains with substantial public health impact but often underfunded relative to need---suggests that standard NIH disease categorization systems may systematically obscure important research concentrations.

\begin{table}[h!]
\centering
\caption{Hidden Domains from ``Other'' Category Identified by LLMs}
\label{tab:hidden_domains}
\footnotesize
\setlength{\tabcolsep}{4pt}
\begin{tabular}{p{5.4cm}r}
\hline
\textbf{Disease/Domain} & \textbf{Count} \\
\hline
Alcohol use disorder & 25 \\
Chronic pain management & 23 \\
Autoimmune diseases & 23 \\
Autism spectrum disorder & 21 \\
Ophthalmology & 19 \\
Substance use disorder & 18 \\
Kidney transplantation & 18 \\
Drug discovery & 17 \\
Chronic obstructive pulmonary disease & 17 \\
Structural biology & 16 \\
Kidney disease & 15 \\
Inflammatory bowel disease & 15 \\
\hline
\end{tabular}
\end{table}

\section{Discussion}

Our comprehensive portfolio-level analysis answers several critical policy questions posed in the introduction and reveals significant structural misalignments between NIH's stated equity priorities and actual AI funding patterns. AI comprises 15.9\% of the NIH portfolio with a measurable 13.4\% funding premium compared to non-AI projects (Findings 1), demonstrating substantial institutional commitment. However, this overall prevalence masks severe geographic and methodological concentration: three disease areas (cancer, aging, mental health) account for 50.1\% of AI investment, while health disparities research comprises only 5.7\% despite being central to NIH's stated mission. This concentration reflects not inevitable research maturity differences, but rather infrastructure advantages in well-funded domains: imaging-intensive methods comprise 37\% of cancer AI work but only 0.9\% of disparities-focused research (Finding 4). This methodological mismatch suggests that established computational tools are not being systematically adapted to address health equity research questions. Additionally, a pronounced research-to-deployment bottleneck limits clinical translation of AI innovations (Finding 3): 79\% of AI projects remain in research/development stages while only 14.7\% engage in clinical deployment or implementation. This gap represents a critical vulnerability in the translation pipeline. Notably, mental health and addiction research achieve higher implementation rates (14.2\% and 11.6\%) due to mature informatics infrastructure and well-developed service delivery systems; health disparities research, by contrast, lacks this advantage, creating a compounding barrier to translating novel AI methods into population health practice.

Our network analysis (Findings 5-7) demonstrates that AI research collaboration, like the broader research ecosystem, follows a hub-and-spoke pattern with significant equity implications. High-volume institutions (Johns Hopkins, Minnesota, Wake Forest) anchor dense intra-community collaboration, while a small bridge layer (University of Washington, Emory, Washington University) provides critical cross-cluster connectivity. The power-law degree distribution ($\alpha=2.08$) indicates that collaboration opportunities and network influence are not broadly distributed: many institutions have limited collaboration breadth while a minority account for disproportionate connectivity. This network structure means that institutions outside these high-centrality positions may face structural barriers to participating in multi-institutional AI research collaborations. Findings 8 and 9 further demonstrate that automated LLM classification, combined with human expert feedback, enables discovery of previously obscured research patterns. The 23.8 percentage point reduction in ``Other'' disease classifications (from 41.5\% to 17.7\%) recovered important research communities in behavioral health, pain management, and autoimmune diseases. This methodological contribution highlights that standard categorical systems, while valuable for administration, may systematically undercategorize emerging or multidisciplinary research areas. The semantic granularity recovered through human-in-the-loop LLM analysis suggests that future portfolio analyses should systematically employ such methods to improve categorization fidelity and reveal underrepresented research directions.

\bibliographystyle{IEEEtran}
\bibliography{IEEEabrv,references}

\end{document}